\documentclass{article} 
\usepackage{iclr2025_conference,times}

\usepackage{amsmath,amsfonts,bm}

\def\eqref#1{equation~\ref{#1}}

\def\1{\bm{1}}

\def\mX{{\bm{X}}}

\DeclareMathAlphabet{\mathsfit}{\encodingdefault}{\sfdefault}{m}{sl}
\SetMathAlphabet{\mathsfit}{bold}{\encodingdefault}{\sfdefault}{bx}{n}

\usepackage{hyperref}
\usepackage{url}
\usepackage{graphicx}
\usepackage{booktabs}
\usepackage{multirow}
\usepackage{subcaption}
\usepackage{enumitem}
\usepackage{amsmath}   
\usepackage{algorithm} 
\usepackage{algpseudocode} 

\title{MMFNet: Multi-Scale Frequency Masking Neural Network for Multivariate Time Series Forecasting}

\author{Aitian Ma, Dongsheng Luo, Mo Sha  \\
Knight Foundation School of Computing and Information Sciences\\
Florida International University\\
Miami, FL, USA \\
\texttt{\{aima,dluo,msha\}@fiu.edu} \\
}

\iclrfinalcopy 
\begin{document}

\maketitle
\begin{abstract}

Long-term Time Series Forecasting (LTSF) is critical for numerous real-world applications, such as electricity consumption planning, financial forecasting, and disease propagation analysis. LTSF requires capturing long-range dependencies between inputs and outputs, which poses significant challenges due to complex temporal dynamics and high computational demands. While linear models reduce model complexity by employing frequency domain decomposition, current approaches often assume stationarity and filter out high-frequency components that may contain crucial short-term fluctuations. In this paper, we introduce MMFNet, a novel model designed to enhance long-term multivariate forecasting by leveraging a multi-scale masked frequency decomposition approach. MMFNet captures fine, intermediate, and coarse-grained temporal patterns by converting time series into frequency segments at varying scales while employing a learnable mask to filter out irrelevant components adaptively.
Extensive experimentation with benchmark datasets shows that MMFNet not only addresses the limitations of the existing methods but also consistently achieves good performance. Specifically, MMFNet achieves up to $6.0\%$ reductions in the Mean Squared Error (MSE) compared to state-of-the-art models designed for multivariate forecasting tasks.

\end{abstract}

\section{Introduction}

Time series forecasting is pivotal in a wide range of domains, such as environmental monitoring~\citep{bhandari2017time}, electrical grid management~\citep{zufferey2017forecasting}, financial analysis~\citep{sezer2020financial}, and healthcare~\citep{zeroual2020deep}. Accurate long-term forecasting is essential for informed decision-making and strategic planning. Traditional methods, such as autoregressive (AR) models~\citep{nassar2004modeling}, exponential smoothing~\citep{hyndman2008forecasting}, and structural time series models~\citep{harvey1989forecasting}, have provided a robust foundation for time series analysis by leveraging historical data to predict future values. However, real-world systems frequently exhibit complex, non-stationary behavior, with time series characterized by intricate patterns such as trends, fluctuations, and cycles. Those complexities pose significant challenges to achieving accurate forecasts~\citep{makridakis1998statistical, box2015time}.

Long-term Time Series Forecasting (LTSF) has seen significant advancements in recent years, driven by the development of sophisticated models, such as Transformer-based models~\citep{zhou2021informer,wu2021autoformer,nie2022time} and linear models~\citep{zeng2023transformers,xu2023fits,lin2024sparsetsf}. Transformer-based architectures have demonstrated exceptional capacity in capturing complex temporal patterns by effectively modeling long-range dependencies through self-attention mechanisms at the cost of heavy computation workload, particularly when facing large-scale time series data, which significantly limits their practicality in real-time applications.

In contrast, the linear models provide a lightweight alternative for real-time forecasting. In particular, FITS demonstrates superior predictive performance across a wide range of scenarios with only $10K$ parameters by utilizing a single-scale frequency domain decomposition method combined with a low-pass filter employing a fixed cutoff frequency~\citep{xu2023fits}. While single-scale frequency domain decomposition offers a global perspective of time series data in the frequency domain, it lacks the ability to localize specific frequency components within the sequence. Furthermore, such methods assume that frequency components remain constant throughout the entire sequence, thereby failing to account for the non-stationary behavior frequently observed in real-world time series. Additionally, the low-pass filter employed by FITS may inadvertently smooth out crucial short-term fluctuations necessary for accurate predictions. The fixed cutoff frequency of the low-pass filter may not be universally optimal for diverse time series datasets, further limiting its adaptability.

In this paper, we present MMFNet, a novel model designed to enhance LTSF through a multi-scale masked frequency decomposition approach. MMFNet captures fine, intermediate, and coarse-grained patterns in the frequency domain by segmenting the time series at multiple scales. At each scale, MMFNet employs a learnable mask that adaptively filters out irrelevant frequency components based on the segment's spectral characteristics.
MMFNet offers two key advantages: (i) the multi-scale frequency decomposition enables MMFNet to effectively capture both short-term fluctuations and broader trends in the data, and (ii) the learnable frequency mask adaptively filters irrelevant frequency components, allowing the model to focus on the most informative signals. These features make MMFNet well-suited to capturing both short-term and long-term dependencies in complex time series, positioning it as an effective solution for various LTSF tasks.

In summary, the contributions of this paper are as follows: \begin{itemize}[leftmargin=*] 
\item To our knowledge, MMFNet is the first model that employs multi-scale frequency domain decomposition to capture the dynamic variations in the frequency domain;
\item MMFNet introduces a novel learnable masking mechanism that adaptively filters out irrelevant frequency components;
\item Extensive experiments show that MMFNet consistently achieves good performance in a variety of multivariate time series forecasting tasks, with up to a $6.0\%$ reduction in the Mean Squared Error (MSE) compared to the existing models. 
\end{itemize}

\section{Preliminaries}

\paragraph{Long-term Time Series Forecasting.} 
LTSF involves predicting future values over an extended time horizon based on previously observed multivariate time series data. The LTSF problem can be formulated as:
\begin{equation}
\hat{x}_{t+1:t+H} = f(x_{t-L+1:t}),
\end{equation}
where \(x_{t-L+1:t} \in \mathbb{R}^{L \times C}\) denotes the historical observation window, and \(\hat{x}_{t+1:t+H} \in \mathbb{R}^{H \times C}\) represents the predicted future values. In this formulation, \(L\) is the length of the historical window, \(H\) is the forecast horizon, and \(C\) denotes the number of features or channels. As the forecast horizon \(H\) increases, the models face challenges to accurately capture both long-term and short-term dependencies within the time series.

\paragraph{Single-Scale Frequency Transformation (SFT).}  
SFT refers to the process of converting the time-domain data into the frequency domain at a single, global scale without segmenting the time series. Such a transformation is typically performed using methods, such as the Fast Fourier Transform (FFT), which efficiently computes the Discrete Fourier Transform (DFT). SFT decomposes the entire signal into sinusoidal components, enabling the analysis of its frequency content. Each frequency component can be expressed as:
\begin{equation}
    X_k = |X_k| e^{j \phi_k},
\end{equation}
where \(|X_k|\) represents the amplitude and \(\phi_k\) the phase of the \(k\)-th frequency component. While the frequency decomposition provides valuable insights into periodic patterns and trends, traditional approaches assume stationarity and operate on a global scale, limiting their capacity to capture the complex, non-stationary characteristics frequently observed in real-world time series. Current frequency-based LTSF models, such as FITS~\citep{xu2023fits}, implement this method by performing frequency domain interpolation at a single scale, which can be formulated as:
\begin{equation}
\tilde{x}_{t+1:t+H} = g(\mathcal{F}(x_{t-L+1:t})),
\end{equation}
where \(\mathcal{F}\) denotes the Fourier transform, and \(g\) represents the filtering operation applied uniformly across the signal. Although SFT is capable of capturing broad temporal patterns, such as long-term trends through low-pass filtering or short-term fluctuations through high-pass filtering, its global application treats the entire signal uniformly. This uniform treatment may result in the loss of important local temporal variations and non-stationary behaviors occurring at different scales.

\section{Method}

\subsection{Overview}\label{sec:mmft}

To overcome the limitations of SFT, we propose the Multi-scale Masked Frequency Transformation (MMFT). MMFT performs frequency decomposition across multiple temporal scales, enabling the model to capture both global and local temporal patterns. Formally, the MMFT problem can be expressed as:
\begin{equation}
\tilde{x}_{t+1:t+H} = h(\{\mathcal{F}_s(x_{t-L+1:t})\}_{s=1}^S),
\end{equation}
where \(\mathcal{F}_s\) denotes the frequency transformation at scale \(s\), and \(h\) represents the aggregation and filtering operation applied to the learnable frequency masks at various scales. Unlike SFT, which applies a single transformation to the entire time series, MMFT divides the signal into multiple scales, each subjected to frequency decomposition. At each scale, a learnable frequency mask is applied to retain the most informative frequency components while selectively discarding noise.
This multi-scale approach allows the model to adapt to non-stationary signals, capturing complex dependencies that span different temporal ranges. By leveraging frequency decomposition at multiple scales and applying adaptive masks, MMFT enhances long-term forecasting accuracy by focusing on both short-term fluctuations and long-term trends within the data. This method increases the model's flexibility and robustness, particularly for non-stationary and multivariate time series. Further analysis of the differences between SFT and MMFT can be found in Appendix~\ref{sec:mmft-sft}.

\begin{figure}[h]
\begin{center}
\includegraphics[width=\textwidth]{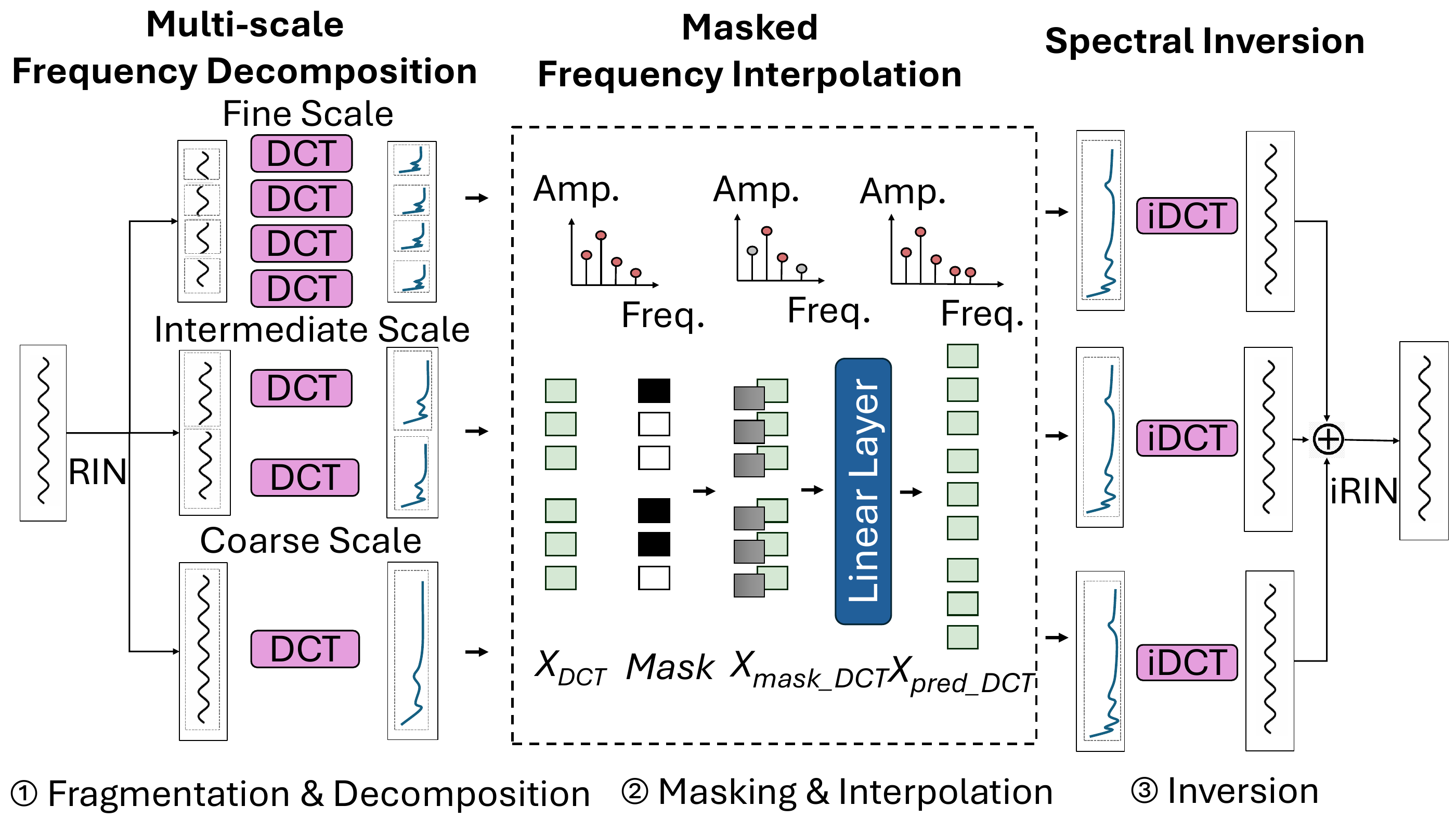}
\end{center}

\caption{MMFNet Architecture. MMFNet consists of the following key components: \textcircled{1} The input time series is first normalized to have zero mean using Reversible Instance-wise Normalization (RIN)~\citep{lai2021revisiting}. The multi-scale frequency decomposition process then divides the time series instance \(X\) into fine, intermediate, and coarse-scale segments, which are subsequently transformed into the frequency domain via the Discrete Cosine Transform (DCT). \textcircled{2} A learnable mask is applied to the frequency segments, followed by a linear layer that predicts the transformed frequency components. \textcircled{3} Finally, the predicted frequency segments from each scale are transformed back into the time domain, merged, and denormalized using inverse RIN (iRIN).
}
\label{arch}
\end{figure}

MMFNet enhances time series forecasting by incorporating the proposed MMFT method to capture intricate frequency features across different scales. The overall architecture of MMFNet is depicted in Figure~\ref{arch}. The model comprises three key components: Multi-scale Frequency Decomposition, Masked Frequency Interpolation, and Spectral Inversion. Multi-scale Frequency Decomposition normalizes the input time series, divides it into segments of varying scales, and transforms these segments into the frequency domain using the DCT. Masked Frequency Interpolation applies a self-adaptive, learnable mask to filter out irrelevant frequency components, followed by a linear transformation of the filtered frequency domain segments. Finally, Spectral Inversion converts the processed frequency components back into the time domain via the Inverse Discrete Cosine Transform (iDCT)~\citep{ahmed1974discrete}. The outputs from different scales are then aggregated, resulting in a refined signal that preserves the essential characteristics of the original input.

\subsection{Multi-scale Frequency Decomposition}

The core concept of Multi-scale Frequency Decomposition lies in applying frequency domain transformations to time series sequences at multiple scales. This approach enables the model to capture both global patterns and fine-grained temporal dynamics by analyzing the data across various segment levels. Multi-scale Frequency Decomposition consists of two fundamental steps: fragmentation and decomposition. Details about the overall workflow can be seen in Appendix~\ref{sec:workflow}.

\paragraph{Fragmentation.}

This step decomposes the time series data into segments of varying lengths to capture features across multiple scales. Specifically, the input sequence \(X\) is first normalized using RIN~\citep{lai2021revisiting} and then partitioned into three sets of segments: fine-scale, intermediate-scale, and coarse-scale segments. Fine-scale segments (\(\mX^{fine}\)) consist of shorter segments that capture detailed, high-frequency components of the time series, enabling the detection of intricate patterns and anomalies that may be missed in longer segments. Intermediate-scale segments (\(\mX^{intermediate}\)) are of moderate length and are designed to capture intermediate-level patterns and trends, striking a balance between the fine and coarse segments. Coarse-scale segments (\(\mX^{coarse}\)) comprise longer segments that capture broader, low-frequency trends and overarching patterns within the data. This multi-scale fragmentation allows the model to effectively capture and leverage patterns across different temporal scales.

\paragraph{Decomposition.}

This step converts the multi-scale time-domain segments into their corresponding frequency components to capture frequency patterns across various temporal scales. For each segment, the DCT is applied to extract frequency domain representations. Specifically, the fine-scale segments in \(\mX^{fine}\) are transformed into \(\mX^{fine}_{DCT}\), the intermediate-scale segments in \(\mX^{intermediate}\) are converted into \(\mX^{intermediate}_{DCT}\), and the coarse-scale segments in \(\mX^{coarse}\) are transformed into \(\mX^{coarse}_{DCT}\).

The DCT for each segment is computed using the following formula:
\begin{equation}
X_{k} = \sum_{n=0}^{N-1} x_{n} \cos \left( \frac{\pi}{N} \left(n + \frac{1}{2}\right) k \right),
\label{eq:dct}
\end{equation}
where \(x_{n}\) represents the time-domain signal values, \(N\) is the segment length, and \(k\) denotes the frequency component. The resulting coefficients \(X_{k}\) represent the frequency components of the segment. This transformation enables MMFNet to capture and analyze patterns at multiple temporal scales in the frequency domain, thereby enhancing its ability to recognize and interpret complex patterns in time series data.

\subsection{Masked Frequency Interpolation}

Masked Frequency Interpolation leverages a learnable mask to adaptively filter frequency components across different scales in the frequency domain, followed by reconstruction through a linear layer neural network. This approach enables the model to learn and apply scale-specific filtering strategies tailored to diverse datasets. The process consists of two primary steps: Masking and Interpolation.

\paragraph{Masking.}

Traditional methods often employ fixed low-pass filters with a predefined cutoff frequency to filter frequency components. These approaches assume that certain frequencies are universally important or irrelevant across the entire time series, an assumption that may not hold for non-stationary data where the relevance of frequency components varies over time. Moreover, over-filtering can lead to the loss of critical details, resulting in oversimplified representations and diminished model performance in tasks such as forecasting and signal analysis. To address these limitations, MMFNet employs an adaptive masking technique to capture dynamic behaviors in the frequency domain. Given the frequency segments \(\mX_{DCT}\), a learnable mask is generated to adaptively filter the frequency components. The mask adjusts the significance of different frequency components by attenuating or emphasizing them based on their relevance to the task. This filtering process is applied via element-wise multiplication, represented as:
\begin{equation}
\mX_{mask\_DCT} = \mX_{DCT} \odot M,
\end{equation}
where \(\odot\) denotes element-wise multiplication, \(M\) represents the learnable mask, and \(\mX_{mask\_DCT}\) is the resulting masked frequency representation. During training, the mask is iteratively updated based on the loss function, allowing MMFNet to focus on the most relevant aspects of the frequency domain representation. This adaptive mechanism improves the model's capacity to capture meaningful patterns while minimizing the influence of irrelevant or noisy information.

\paragraph{Interpolation.}

In this step, the masked frequency segments \(\mX_{mask\_DCT}\) are transformed into predicted frequency domain segments \(\mX_{pred\_DCT}\) through a linear layer. This linear transformation maps the filtered frequency components to the target frequency representations aligned with the model's forecasting objectives. Specifically, a fully connected (dense) layer is applied to the masked frequency components, and this operation can be expressed as:
\begin{equation} 
\mX_{pred\_DCT} = W \cdot \mX_{mask\_DCT} + b,
\end{equation}
where \(W\) denotes the weight matrix of the linear layer, and \(b\) is the bias term. The linear layer is designed to learn a projection that aligns the filtered frequency components with the target prediction space. This transformation further refines the frequency domain information, producing \(\mX_{pred\_DCT}\), which is essential for reconstructing accurate time-domain predictions. By leveraging the refined frequency information and reducing the influence of irrelevant frequency components, this step improves the overall prediction accuracy.

\subsection{Spectral Inversion}

The final process, Spectral Inversion, transforms the interpolated frequency components back into the time domain using the iDCT, reversing the earlier DCT process. The iDCT is applied individually to the predicted frequency domain segments \(\mX^{fine}_{pred\_DCT}\), \(\mX^{intermediate}_{pred\_DCT}\), and \(\mX^{coarse}_{pred\_DCT}\). The iDCT for a segment is given by the following formula:
\begin{equation}
x_{n} = \frac{1}{2} x_{0} + \sum_{k=1}^{N-1} X_{k} \cos \left( \frac{\pi}{N} \left(n + \frac{1}{2}\right) k \right),
\label{eq:idct}
\end{equation}
where \(x_{n}\) represents the time-domain signal values, \(X_{k}\) are the frequency components, and \(N\) denotes the segment length. This equation reconstructs the time-domain signal by summing the contributions of each frequency component~\citep{davis1984discrete}.

Once the iDCT is performed separately for each scale, the resulting time-domain signals are combined. This integration step merges the multi-scale frequency information by combining the outputs from the fine, intermediate, and coarse scales. The final reconstructed signal preserves the key characteristics of the original input while incorporating the enhanced interpolation achieved through the masked frequency filtering.

\section{Experiment}

In this section, we evaluate MMFNet with several LTSF benchmark datasets across a range of forecast horizons. We also conduct ablation studies to assess the impact of MMFT and our frequency masking techniques. Finally, we evaluate MMFNet’s performance in ultra-long-term forecasting scenarios.

\subsection{Experimental Setup}

\paragraph{Datasets.}
We perform experiments with seven widely-used LTSF datasets: ETTh1, ETTh2, ETTm1, ETTm2, Weather, Electricity, and Traffic. More details on those datasets can be found in Appendix~\ref{sec:datasets}.

\paragraph{Baselines.}
We compare MMFNet against several state-of-the-art models, including FEDformer~\citep{zhou2022fedformer}, TimesNet~\citep{wu2022timesnet}, TimeMixer~\citep{wang2024timemixer}, and PatchTST~\citep{nie2022time}. In addition, we compare MMFNet against several lightweight models, including DLinear~\citep{zeng2023transformers}, FITS~\citep{xu2023fits}, and SparseTSF~\citep{lin2024sparsetsf}. More details on our baseline models can be found in Appendix~\ref{sec:baselines}.

\paragraph{Environment.}
All experiments are implemented using PyTorch~\citep{paszke2019pytorch} and run on a single NVIDIA GeForce RTX 4090 GPU with $24$GB of memory.

\subsection{Performance on LTSF Benchmarks}

\begin{table}[ht]
    \centering
    \caption{Multivariate LTSF MSE results on ETT, Weather, Electricity, and Traffic. The best result is emphasized in \textbf{bold}, while the second-best is \underline{underlined}. ``Imp.'' represents the improvement between MMFNet and either the best or second-best result, with a higher ``Imp.'' indicating greater improvement.}
    \label{tab:mse_results}
    \resizebox{\textwidth}{!}{
    \begin{tabular}{ccccccccccc}

        \toprule

 \multicolumn{2}{c}{Models} & MMFNet & FITS & SparseTSF & DLinear & PatchTST & TimeMixer & TimesNet & FEDformer & \multirow{2}{*}{Imp.}\\
        \cmidrule{1-2}
               Data & Horizon &  (ours) &  (2024) &  (2024) &  (2023) &  (2023) &  (2024) &  (2023) &  (2022) & \\

        \midrule
       \multirow{4}{*}{\rotatebox{90}{ETTh1}} &96  & \textbf{0.359} & 0.372 & \underline{0.362} & 0.384 & 0.385 & 0.380 & 0.384 & 0.375 &+0.003 \\
         &192 & \textbf{0.396} & 0.404 & \underline{0.403} & 0.443 & 0.413 & 0.413 & 0.436 & 0.427 &+0.006 \\
         &336 & \textbf{0.409} & \underline{0.427} & 0.434 & 0.446 & 0.440 & 0.445 & 0.491 & 0.459 &+0.018 \\
         &720 & \textbf{0.419} & \underline{0.424} & 0.426 & 0.504 & 0.456 & 0.491 & 0.521 & 0.484 &+0.005 \\
        
        \midrule
        \multirow{4}{*}{\rotatebox{90}{ETTh2}}&96  & \textbf{0.263} & \underline{0.271} & 0.294 & 0.282 & 0.274 & 0.281 & 0.340 & 0.340 & +0.008 \\
        &192 & \textbf{0.317} & \underline{0.331} & 0.339 & 0.340 & 0.338 & 0.356 & 0.402 & 0.433 &+0.014 \\
        &336 & \textbf{0.336} & \underline{0.354} & 0.359 & 0.414 & 0.367 & 0.371 & 0.452 & 0.508 &+0.018 \\
        &720 & \textbf{0.376} & \underline{0.377} & 0.383 & 0.588 & 0.391 & 0.403 & 0.462 & 0.480 &+0.001 \\
        
        \midrule
        \multirow{4}{*}{\rotatebox{90}{ETTm1}}&96  & 0.307 & 0.303 & 0.314 & \underline{0.301} & \textbf{0.292} & 0.315 & 0.338 & 0.362&-0.015 \\
        &192 & \underline{0.334} & 0.337 & 0.343 & 0.335 & \textbf{0.330} & 0.339 & 0.374 & 0.393&-0.004 \\
        &336 & \textbf{0.358} & 0.366 & 0.369 & 0.371 & \underline{0.365} & 0.366 & 0.410 & 0.442&+0.007 \\
        &720 & \textbf{0.396} & \underline{0.415} & 0.418 & 0.426 & 0.419 & 0.423 & 0.478 & 0.483&+0.019 \\
         
        \midrule
        \multirow{4}{*}{\rotatebox{90}{ETTm2}}&96  & \textbf{0.160} & \underline{0.162} & 0.165 & 0.171 & 0.163 & 0.176 & 0.187 & 0.189&+0.002 \\
        &192 & \textbf{0.212} & \underline{0.216} & 0.218 & 0.237 & 0.219 & 0.226 & 0.249 & 0.256&+0.004 \\
        &336 & \textbf{0.259} & \underline{0.268} & 0.272 & 0.294 & 0.276 & 0.276 & 0.321 & 0.326&+0.009 \\
        &720 & \textbf{0.327} & \underline{0.348} & 0.352 & 0.426 & 0.368 & 0.372 & 0.408 & 0.437&+0.021 \\
        
        \midrule
        \multirow{4}{*}{\rotatebox{90}{Weather}}&96  & 0.153 & \textbf{0.143} & 0.172 & 0.174 & \underline{0.151} & 0.159 & 0.172 & 0.246&-0.010 \\
        &192 & \underline{0.194} & \textbf{0.186} & 0.215 & 0.217 & 0.195 & 0.202 & 0.219 & 0.292&-0.008 \\
        &336 & \underline{0.241} & \textbf{0.236} & 0.263 & 0.262 & 0.249 & 0.281 & 0.280 & 0.378&-0.005 \\
        &720 & \textbf{0.302} & \underline{0.307} & 0.318 & 0.332 & 0.321 & 0.335 & 0.365 & 0.447&+0.005 \\
        
        \midrule
        \multirow{4}{*}{\rotatebox{90}{Electricity}} &96  & \underline{0.131} & 0.134 & 0.138 & 0.140 & \textbf{0.129} & 0.158 & 0.168 & 0.188 &-0.002 \\
        &192 & \textbf{0.146} & \underline{0.149} & 0.151 & 0.153 & \underline{0.149} & 0.174 & 0.184 & 0.197 &+0.003 \\
        &336 & \textbf{0.162} & \underline{0.165} & 0.166 & 0.169 & 0.166 & 0.190 & 0.198 & 0.212&+0.003 \\
        &720 & \textbf{0.199} & \underline{0.203} & 0.205 & 0.204 & 0.210 & 0.229 & 0.220 & 0.244&+0.004 \\
        
        \midrule
        \multirow{4}{*}{\rotatebox{90}{Traffic}} &96  & 0.381 & 0.385 & 0.389 & 0.413 & \textbf{0.366} & \underline{0.380} & 0.593 & 0.573&-0.015 \\
        &192 & \underline{0.394} & 0.397 & 0.398 & 0.423 & \textbf{0.388} & 0.397 & 0.617 & 0.611&-0.006 \\
        &336 & \underline{0.408} & 0.410 & 0.411 & 0.437 & \textbf{0.398} & 0.418 & 0.629 & 0.621&-0.010 \\
        &720 & \underline{0.446} & 0.448 & 0.448 & 0.466 & 0.457 & \textbf{0.436} & 0.640 & 0.630&-0.010 \\
        \bottomrule
    \end{tabular}}
\end{table}

The experimental results offer several key insights into MMFNet's performance across a range of datasets and forecast horizons. As Table~\ref{tab:mse_results} shows, MMFNet demonstrates superior performance on the ETT dataset and consistently achieves the best results even at extended forecasting horizons. Additionally, it maintains strong performance across a range of channel numbers and sampling rates.

\paragraph{Performance on the ETT Dataset.} 
As Table~\ref{tab:mse_results} shows, MMFNet consistently outperforms other models across all forecast horizons on the ETTh1, ETTh2, and ETTm2 datasets. 
For example, on ETTh1, compared with other baseline models, MMFNet achieves the best MSE results of $0.359$, $0.396$, $0.409$, and $0.419$ at forecast horizons of $96$, $192$, $336$, and $720$, respectively. Moreover, it demonstrates a $4.2\%$ MSE reduction (+0.018) at the forecast horizon of $336$ on ETTh1 and a $5.1\%$ MSE reduction (+0.018) at the forecast horizon of $336$ on ETTh2. This consistent performance highlights MMFNet's ability to effectively capture both short-term fluctuations and long-term dependencies in time series data, positioning it as a versatile model for a wide variety of LTSF tasks.

\paragraph{Performance at the Extended Horizon.} 
As Table~\ref{tab:mse_results} shows, at the extended forecast horizon of $720$, MMFNet consistently achieves the highest predictive accuracy across all datasets, except for Traffic where it ranks second. Notably, MMFNet demonstrates significant improvements over baseline models, achieving MSE reductions of $4.6\%$ ($+0.019$) on ETTm1 and $6.0\%$ ($+0.021$) on ETTm2 at forecast horizon $720$ compared to the second-best models. These results highlight the robustness of MMFNet in addressing long-term forecasting tasks.

\paragraph{Performance in Low-Channel, Low-Sampling Rate Scenarios.} 
As Table~\ref{tab:mse_results} shows, in scenarios involving datasets with fewer channels (7 channels) and lower sampling rates (1-hour intervals), such as in the ETTh1 and ETTh2 datasets, linear models like FITS, SparseTSF, and DLinear exhibit strong performance. For example, on ETTh2, FITS achieves the MSE results of $0.271$, $0.331$, $0.354$, and $0.377$ at forecast horizons of $96$, $192$, $336$, and $720$, respectively. MMFNet continues to surpass these models on ETTh2 by achieving the MSE results of $0.263$, $0.317$, $0.336$, and $0.376$ at forecast horizons of $96$, $192$, $336$, and $720$, respectively. This suggests that multi-scale frequency decomposition methods are particularly well-suited for datasets with fewer channels and broader time intervals between measurements.

\paragraph{Performance in High-Channel Scenarios.} 
As Table~\ref{tab:mse_results} shows, for datasets with larger numbers of channels, such as Electricity ($321$ channels, $1$-hour sampling rate) and Traffic ($862$ channels, $1$-hour sampling rate), MMFNet and FITS consistently demonstrate strong performance. Despite the increased complexity that arises from higher channel counts.
For example, on Electricity,  MMFnet achieves the best MSE results of $0.146$, $0.162$, and $0.199$ at forecast horizons of $192$, $336$, and $720$, respectively.
MMFNet's multi-scale frequency decomposition enables it to effectively model complex temporal dependencies while maintaining high predictive accuracy. While PatchTST performs better on the traffic dataset, it leverages a patching transformer mechanism rather than a purely linear frequency-based approach, distinguishing it from MMFNet and FITS in terms of the model architecture. This further indicates that more sophisticated decomposition methods are required for lightweight models to handle high-channel scenarios effectively.

\paragraph{Performance in High-Sampling Rate Scenarios.} 

As Table~\ref{tab:mse_results} shows, for datasets with higher sampling rates, such as Weather ($21$ channels, $10$-minute sampling rate), ETTm1 and ETTm2 ($7$ channels, $15$-minute sampling rate), MMFNet and FITS consistently demonstrate strong performance. For example, on ETTm2,  MMFnet achieves the best MSE results of $0.160$, $0.212$, $0.259$, and $0.327$ at forecast horizons of $96$, $192$, $336$, and $720$, respectively. Despite the increased complexity that arises from a faster sampling rate, MMFNet's multi-scale frequency decomposition enables it to effectively model complex temporal dependencies while maintaining high predictive accuracy.

\subsection{Comparisons between MMFT and SFT}

\begin{table}[ht]
    \centering

    \caption{MSE values of MMFNet when it uses SFT and MMFT on the ETT dataset. SFT denotes the standard single-scale frequency decomposition approach. MFT refers to the masked frequency transformation with fragmentation applied at a single scale, where \(N_{seg}\) specifies the segment length. MMFT denotes the full MMFT method, which performs frequency decomposition with multi-scale fragmentation. ``Imp." indicates the improvement of MMFT over SFT.}
    \label{tab:seg_len}
    \resizebox{0.9\textwidth}{!}{
    \begin{tabular}{lcccccccccccccccccc}
        \toprule

        Dataset & & \multicolumn{4}{c}{ETTh1} & \multicolumn{4}{c}{ETTh2} \\
        \cmidrule(lr){1-2} \cmidrule(lr){3-6} \cmidrule(lr){7-10} 
        Horizon& & 96 & 192 & 336 & 720 & 96 & 192 & 336 & 720   \\
        \midrule

        SFT   & & 0.372 & 0.404 & 0.427 & 0.424 & 0.271 & 0.331 & 0.354 & 0.377 \\
 
       MFT ($N_{seg}= 24$) & & 0.362 & 0.400 & 0.412 & 0.421 & 0.264 & 0.317 & 0.336 & 0.376 \\
        
        MFT ($N_{seg}= 120$) & & 0.366 & 0.401 & 0.426 & 0.423 & 0.265 & 0.317 & 0.336 & 0.376  \\
        
        MFT ($N_{seg}= 360$) & & 0.366 & 0.403 & 0.418 & 0.425 & 0.265& 0.317 & 0.340 & 0.376  \\

        MMFT   & & \textbf{0.359} & \textbf{0.396} & \textbf{0.409} & \textbf{0.419} & \textbf{0.263} & \textbf{0.317} & \textbf{0.336} & \textbf{0.376} \\

        \midrule
        Imp.(MMFT over SFT) & & +0.013 & +0.008 & +0.018 & +0.005 & +0.008 & +0.014 & +0.018 & +0.001  \\
  
        \bottomrule
    \end{tabular}
    }
\end{table}

To evaluate the effectiveness of the MMFT method (see Section~\ref{sec:mmft}), we perform experiments using the ETT dataset. Both SFT and MMFT incorporate the same adaptive masking strategy to ensure fair and consistent comparisons. SFT applies FFT to the entire time series without fragmentation, while MFT introduces a single-scale fragmentation, and MMFT performs a multi-scale fragmentation. The results presented in Table~\ref{tab:seg_len} reveal two important insights. 

First, fragmentation consistently enhances frequency domain decomposition. On the ETTh1 dataset, MFT ($N_{seg}= 360$) achives the MSE results of $0.160$, $0.212$, $0.259$, and $0.327$ at forecast horizons of $96$, $192$, $336$, and $720$, respectively. MFT delivers the most significant gains observed at a segment length of $24$ with a $4.2\%$ MSE reduction (+0.018) at then forecast horizon of $336$. This improvement suggests that segmenting the time series into smaller segments enables MFT to capture localized frequency features more effectively.

Second, MMFT, leveraging multi-scale decomposition, consistently delivers superior results compared to both SFT and single-scale MFT. 
On the ETTh2 dataset, MMFT achives the MSE results of $0.263$, $0.317$, $0.336$, and $0.376$ at forecast horizons of $96$, $192$, $336$, and $720$, respectively. At the forecast horizon of $336$, MMFT achieves substantial reductions in MSE, including a $0.018$ improvement over SFT. These results suggest that the multi-scale decomposition employed by MMFT allows for the capture of a broader range of frequency patterns, leading to more accurate predictions, particularly in long-term forecasting scenarios.

\subsection{Effectiveness of Masking}

\begin{table}[ht]
    \centering
    \caption{MSE results for multivariate LTSF with MMFNet on the ETT dataset with or without the masking module. ``Mask'' refers to results with the masking module, while ``w/o Mask'' refers to results without it. ``Imp.'' denotes the improvement enabled by the masking module.}
    \label{tab:mask}
    \resizebox{\textwidth}{!}{
    \begin{tabular}{lcccccccccccccccccc}
        \toprule
        Dataset & & \multicolumn{4}{c}{ETTh1} & \multicolumn{4}{c}{ETTh2} & \multicolumn{4}{c}{Electricity} & \multicolumn{4}{c}{Traffic} \\
        \cmidrule(lr){1-2} \cmidrule(lr){3-6} \cmidrule(lr){7-10} \cmidrule(lr){11-14} \cmidrule(lr){15-18}
        Horizon& & 96 & 192 & 336 & 720 & 96 & 192 & 336 & 720 & 96 & 192 & 336 & 720 & 96 & 192 & 336 & 720 \\
        \midrule
        
        w/o Mask & & 0.372 & 0.405 & 0.410 & 0.420 & 0.269 & 0.319 & 0.339 & 0.376 & 0.312 & 0.338 & 0.360 & 0.397 & 0.166 & 0.218 & 0.264 & 0.330 \\

       Mask & & \textbf{0.359} & \textbf{0.396} & \textbf{0.409} & \textbf{0.419} & \textbf{0.263} & \textbf{0.317} & \textbf{0.336} & \textbf{0.376} & \textbf{0.307} & \textbf{0.334} & \textbf{0.358} & \textbf{0.396} & \textbf{0.160} & \textbf{0.212} & \textbf{0.259} & \textbf{0.327} \\
       
        \midrule
        Imp. & & +0.013 & +0.009 & +0.001 & +0.001 & +0.006 & +0.002 & +0.003 & +0.000 & +0.005 & +0.003 & +0.002 & +0.001 & +0.006 & +0.006 & +0.005 & +0.003 \\
  
        \bottomrule
    \end{tabular}
    }
\end{table}

To evaluate the effectiveness of the self-adaptive masking mechanism, we compare MMFNet's performance on the ETT dataset with and without the masking module across four forecast horizons: $96$, $192$, $336$, and $720$. 
As Table~\ref{tab:mask} lists, MMFNet with masking consistently outperforms the version without masking across all horizons. The most notable improvements occur at the horizon $96$ with a $3.5\%$ MSE reduction on ETTh1 (+0.013) and a $2.2\%$ MSE reduction on ETTh2 (+0.006). With the Electricity dataset, the largest improvement is at horizon $96$ with an improvement of $+0.005$. Similarly, the largest improvement is at horizon $192$ with an improvement of $+0.006$ on the Traffic dataset. The results show that the self-adaptive masking mechanism which filters out frequency noise at different scales consistently enhances forecasting accuracy across various datasets and forecast horizons. A more detailed analysis of the mask output is provided in Appendix~\ref{sec:mask_output}.

\subsection{Performance on Ultra-long-term Time Series Forecasting}\label{sec:ultra-long}

\begin{table}[ht]
    \centering
    \caption{MSE results for multivariate ultra long-term time series forecasting with MMFNet. The best result is emphasized in \textbf{bold}, while the second-best is \underline{underlined}. ``Imp.'' represents the improvement between MMFNet and either the best or second-best result, with a higher ``Imp.'' value indicating greater improvement.}
    \label{tab:long}
    \resizebox{\textwidth}{!}{
    \begin{tabular}{lcccccccccccccccccc}
        \toprule
        Dataset & & \multicolumn{4}{c}{ETTm1} & \multicolumn{4}{c}{ETTm2} & \multicolumn{4}{c}{Electricity} & \multicolumn{4}{c}{Weather} \\
        \cmidrule(lr){1-2} \cmidrule(lr){3-6} \cmidrule(lr){7-10} \cmidrule(lr){11-14} \cmidrule(lr){15-18}
        Horizon& & 960 & 1200 & 1440 & 1680 & 960 & 1200 & 1440 & 1680 & 960 & 1200 & 1440 & 1680 & 960 & 1200 & 1440 & 1680 \\
        \midrule
        
            DLinear & & 0.429 & 0.440 & 0.463 & 0.481 & 0.412 & 0.398 & 0.430 & 0.478 & 0.238 & 0.267 & 0.277 & \underline{0.296} & 0.330 & 0.341 & \underline{0.345} & 0.356 \\

       FITS & & \underline{0.413} & 0.422 & 0.425 & 0.427 & \underline{0.347} & \underline{0.358} & \textbf{0.355} & \underline{0.350} & 0.238 & 0.268 & 0.293 & 0.311 & 0.333 & 0.343 & 0.353 & 0.360 \\

         SparseTSF & &  0.415 & \underline{0.422} & \underline{0.424} & \underline{0.425} & 0.353 & 0.367 & 0.357 & 0.353 & \underline{0.228} & \underline{0.256} & \underline{0.281} & 0.298 & \underline{0.329} & \underline{0.339} & 0.347 & \underline{0.353} \\
       
        \midrule

       MMFNet(ours) & & \textbf{0.411} & \textbf{0.419} & \textbf{0.423} & \textbf{0.424} & \textbf{0.346} & \textbf{0.357} & \underline{0.356} & \textbf{0.349} & \textbf{0.224} & \textbf{0.255} & \textbf{0.280} & \textbf{0.292} & \textbf{0.318} & \textbf{0.331} & \textbf{0.340} & \textbf{0.349} \\
       
        \midrule
        Imp. & & +0.002 & +0.003 & +0.001 & +0.001 & +0.001 & +0.001 & -0.001 & +0.001 & +0.004 & +0.001 & +0.001 & +0.004 & +0.011 & +0.008 & +0.005 & +0.004 \\
  
        \bottomrule
    \end{tabular}
    }
\end{table}

We evaluate MMFNet's performance in ultra-long-term time series forecasting scenarios. Table~\ref{tab:long} presents the MSE results for various models applied to multivariate ultra-long-term time series forecasting across four datasets at forecast horizons of $960$, $1200$, $1440$, and $1680$. Due to the significant memory requirements of models such as FEDformer, TimesNet, TimeMixer, and PatchTST when forecast horizons are extended, these models exceed GPU memory limitations. Consequently, in this context, we limit the comparison to more lightweight models: DLinear, FITS, SparseTSF, and the proposed MMFNet.

The results show that MMFNet consistently outperforms the existing models across most datasets and forecast horizons. For example, with the ETTh1 dataset, MMFNet achieves the MSE values of $0.411$, $0.419$, $0.423$, and $0.424$ at horizons of $960$, $1200$, $1440$, and $1680$, respectively. With the Electricity dataset, MMFNet delivers very good performance, particularly at longer horizons, with the MSE values of $0.255$ at $1200$ and $0.292$ at $1680$.On the Weather dataset, MMFNet demonstrates superior performance, achieving MSE values of $0.318$ at the $960$ horizon and $0.331$ at the $1200$ horizon, representing a $3.3\%$ ($+0.011$) and $2.4\%$ ($+0.008$) reduction in MSE compared to the second-best baseline. The results demonstrate the robustness of MMFNet in forecasting multivariate ultra-long-term time series data across various datasets and extended forecast horizons by effectively capturing frequency variations at different scales.
\section{Related Work}

\subsection{Long-term Time Series Forecasting }

LTSF is a critical area in data science and machine learning and focuses on predicting future values over extended periods. Such a task is challenging due to the inherent seasonality, trends, and noise in time series data. In addition, time series data is often complex and high-dimensional~\cite{zheng2024parametric,zheng2023auto}. Traditional statistical methods, such as ARIMA~\citep{contreras2003arima} and Holt-Winters~\citep{chatfield1988holt}, are effective for short-term forecasting but frequently fall short for longer horizons. Machine learning models, such as SVM~\citep{wang2005comparison}, Random Forests~\cite{breiman2001random}, and Gradient Boosting Machines~\citep{natekin2013gradient}, offer improved performance by capturing non-linear relationships but typically require extensive feature engineering. Recently, deep learning models, such as RNNs, LSTMs, GRUs, and Transformer-based models (Informer and Autoformer), have demonstrated notable efficiency in modeling long-term dependencies. Furthermore, the hybrid models that combine statistical methods with machine learning or deep learning techniques have shown improved accuracy. State-of-the-art models, such as FEDformer~\citep{zhou2022fedformer}, FiLM~\citep{zhou2022film}, PatchTST~\cite{nie2022time}, and SparseTSF, leverage frequency domain transformations and efficient self-attention to improve prediction performance.

\subsection{Time Series Forecasting in the Frequency Domain}\label{sec:freq}

Recent advancements in time series analysis have increasingly utilized frequency domain information to reveal underlying patterns. For instance, FNet~\citep{lee2021fnet} adopts an attention-based approach to capture temporal dependencies within the frequency domain, thereby eliminating the need for convolutional or recurrent layers. Models such as FEDformer~\citep{zhou2022fedformer} and FiLM~\citep{zhou2022film} improve predictive performance by incorporating frequency domain information as auxiliary features. FITS~\citep{xu2023fits} also demonstrates strong predictive capabilities by converting time-domain forecasting tasks into the frequency domain and utilizing low-pass filters to reduce the number of parameters required. However, many of these techniques rely on manual feature engineering to identify dominant periods, which can constrain the amount of information captured and introduce inefficiencies or risks of overfitting.

\subsection{Multiscaling Model}

In the field of computer vision, several multi-scale Vision Transformers (ViTs) have leveraged hierarchical architectures to generate progressively down-sampled pyramid features. For instance, Multi-Scale Vision Transformers~\citep{fan2021multiscale} enhance the standard Vision Transformer architecture by incorporating multi-scale processing, allowing for improved detail capture across varying spatial resolutions. Pyramid Vision Transformer~\citep{wang2021pyramid} integrates a pyramid structure within ViTs to facilitate multi-scale feature extraction, while Twins~\citep{twins} combines local and global attention to effectively model multi-scale representations. SegFormer~\citep{segformer} introduces an efficient hierarchical encoder that captures both coarse and fine features, and CSWin~\citep{dong2022cswin} further improves performance by using multi-scale cross-shaped local attention mechanisms.
In the context of time series forecasting, TimeMixer~\citep{wang2024timemixer} represents a significant advancement with its fully MLP-based architecture, which employs Past-Decomposable-Mixing and Future-Multipredictor-Mixing blocks. This architecture enables TimeMixer to effectively leverage disentangled multi-scale time series data during both past extraction and future prediction phases.

\subsection{Masked Modeling}

Masked language modeling and its autoregressive variants have emerged as dominant self-supervised learning approaches in natural language processing. These techniques enable large-scale language models to excel in both language understanding and generation by predicting masked or hidden tokens within sentences~\citep{devlin2018bert, radford2018improving}. In computer vision, early approaches, such as the context encoder~\citep{pathak2016context}, involve masking specific regions of an image and predicting the missing pixels, while Contrastive Predictive Coding~\citep{oord2018representation} uses contrastive learning to improve feature representations. Recent innovations in MIM include models like iGPT~\citep{chen2020generative}, ViT~\citep{dosovitskiy2020image}, and BEiT~\citep{bao2021beit}, which leverage Vision Transformers and techniques, such as pixel clustering, mean color prediction, and block-wise masking. In the realm of multivariate time series forecasting, masked encoders have recently been employed with notable success in classification and regression tasks~\citep{zerveas2021transformer}. For example, PatchTST uses a masked self-supervised representation learning method to reconstruct the masked patches and showcases its effectiveness in time series data~\citep{nie2022time}. However, the application of masked modeling techniques in linear time series forecasting remains relatively under-explored.

\section{Conclusion}

MMFNet significantly advances long-term multivariate forecasting by employing the MMFT approach. Through comprehensive evaluations on benchmark datasets, we have demonstrated that MMFNet consistently outperforms state-of-the-art models in forecasting accuracy, highlighting its robustness in capturing complex data patterns. By effectively integrating multi-scale decomposition with a learnable masked filter, MMFNet captures intricate temporal details while adaptively mitigating noise, making it a versatile and reliable solution for a wide range of LTSF tasks.

\bibliography{MMFNet}
\bibliographystyle{iclr2025_conference}

\appendix

\section{More on MMFNet}\label{sec:mmfnet}

\subsection{Overall Workflow}\label{sec:workflow}

The overall workflow of MMFNet is presented in Algorithm~\ref{alg:mmfnet}. The algorithm takes a univariate historical look-back window as input, \(x_{t-L+1:t}\), and produces the corresponding forecast, \(\hat{x}_{t+1:t+H}\). By incorporating the channel-independent strategy, in which multiple channels are modeled using a shared set of parameters, MMFNet can efficiently extend to multivariate time series forecasting tasks. Such an approach enables the model to leverage its multi-scale frequency decomposition and adaptive masking framework across various input channels to enhance its predictive capabilities in complex multivariate settings.

\begin{algorithm}
\caption{Overall Pseudocode of MMFNet}
\label{alg:mmfnet}
\begin{algorithmic}[1]

\Require 
    Historical look-back window $x_{t-L+1:t} \in \mathbb{R}^{L }$
\Ensure 
    Forecasted output $\hat{x}_{t+1:t+H} \in \mathbb{R}^{H}$

\State $x_d \leftarrow \text{RIN}(x_{t-L+1:t})$ 
    \Comment{Apply Reversible Instance-wise Normalization (RIN)}

\State $X_{\text{fine}} \leftarrow \text{Reshape}(x_d,(n_{fine},s_{fine}))$ 
    \Comment{Reshape $x_d$ into a $n_{fine} \times s_{fine}$ matrix}

\State $X_{DCT}^{\text{fine}} \leftarrow \text{DCT}(X_{\text{fine}})$ 
    \Comment{Apply DCT to each segment with Equation~\ref{eq:dct}}

\State $X^{\text{fine}}_{\text{mask\_DCT}} \leftarrow X^{\text{fine}}_{\text{DCT}} \odot \text{Mask}_{fine}$
    \Comment{Apply the learnable mask}

\State $x^{\text{fine}}_{\text{mask\_DCT}} \leftarrow  \text{Reshape}(X^{\text{fine}}_{\text{mask\_DCT}})$
    \Comment{Reshape the matrix back to a sequence of length $L$}

\State $x^{\text{fine}}_{\text{pred\_DCT}} \leftarrow \text{Linear}(x^{\text{fine}}_{\text{mask\_DCT}})$
    \Comment{Apply a linear transformation}

\State $x_{\text{fine\_pred}} \leftarrow \text{iDCT}(x^{\text{fine}}_{\text{pred\_DCT}})$
    \Comment{Apply iDCT to recover the time domain with Equation~\ref{eq:idct}}

\State $X_{\text{inter}} \leftarrow \text{Reshape}(x_d,(n_{inter},s_{inter}))$ 
    \Comment{Reshape $x_d$ into a $n_{inter} \times s_{inter}$ matrix}

\State $X_{DCT}^{\text{inter}} \leftarrow \text{DCT}(X_{\text{inter}})$ 
    \Comment{Apply DCT to each intermediate-scale segment  with Equation~\ref{eq:dct}}

\State $X^{\text{inter}}_{\text{mask\_DCT}} \leftarrow X^{\text{inter}}_{\text{DCT}} \odot \text{Mask}_{inter}$
    \Comment{Apply the learnable mask}

\State $x^{\text{inter}}_{\text{mask\_DCT}} \leftarrow  \text{Reshape}(X^{\text{inter}}_{\text{mask\_DCT}})$
    \Comment{Reshape the matrix back to a sequence of length $L$}

\State $x^{\text{inter}}_{\text{pred\_DCT}} \leftarrow \text{Linear}(x^{\text{inter}}_{\text{mask\_DCT}})$
    \Comment{Apply a linear transformation}

\State $x_{\text{inter\_pred}} \leftarrow \text{iDCT}(x^{\text{inter}}_{\text{pred\_DCT}})$
    \Comment{Apply iDCT to recover the time domain with Equation~\ref{eq:idct}}

\State $X_{\text{coarse}} \leftarrow \text{Reshape}(x_d,(n_{coarse},s_{coarse}))$ 
    \Comment{Reshape $x_d$ into a $n_{coarse} \times s_{coarse}$ matrix}

\State $X_{DCT}^{\text{coarse}} \leftarrow \text{DCT}(X_{\text{coarse}})$ 
    \Comment{Apply DCT to each coarse-scale segment with Equation~\ref{eq:dct}}

\State $X^{\text{coarse}}_{\text{mask\_DCT}} \leftarrow X^{\text{coarse}}_{\text{DCT}} \odot \text{Mask}_{coarse}$
    \Comment{Apply the learnable mask}

\State $x^{\text{coarse}}_{\text{mask\_DCT}} \leftarrow  \text{Reshape}(X^{\text{coarse}}_{\text{mask\_DCT}})$
    \Comment{Reshape the matrix back to a sequence of length $L$}

\State $x^{\text{coarse}}_{\text{pred\_DCT}} \leftarrow \text{Linear}(x^{\text{coarse}}_{\text{mask\_DCT}})$
    \Comment{Apply a linear transformation}

\State $x_{\text{coarse\_pred}} \leftarrow \text{iDCT}(x^{\text{coarse}}_{\text{pred\_DCT}})$
    \Comment{Apply iDCT to recover the time domain with Equation~\ref{eq:idct}}

\State $x_M \leftarrow x_{\text{fine\_pred}} + x_{\text{inter\_pred}} + x_{\text{coarse\_pred}} + e_t$
    \Comment{Combine predictions from all scales and add back the mean}

\State $\hat{x}_{t+1:t+H} \leftarrow \text{iRIN}(x_M)$
    \Comment{Apply inverse Reversible Instance-wise Normalization (iRIN)}

\end{algorithmic}
\end{algorithm}

\subsection{Detailed Dataset Description}\label{sec:datasets}

\begin{table}[htbp]
\centering
\caption{Statistics of the datasets.}
\label{tab:dataset_statistics}
    \resizebox{0.8\textwidth}{!}{

\begin{tabular}{c|c|c|c|c|c|c|c}
        \toprule

Dataset & Traffic & Electricity & Weather & ETTh1 & ETTh2 & ETTm1 & ETTm2 \\ 
        \midrule

Channels & 862 & 321 & 21 & 7 & 7 & 7 & 7 \\ 
        \midrule

Sampling Rate & 1 hour & 1 hour & 10 min & 1 hour & 1 hour & 15 min & 15 min \\ 
        \midrule

Total Timesteps & 17,544 & 26,304 & 52,696 & 17,420 & 17,420 & 69,680 & 69,680 \\ 

\bottomrule
\end{tabular}
}
\end{table}

Here is a brief description of the datasets used in our experiments. 

\begin{itemize}
    \item The ETT dataset\footnote{https://github.com/zhouhaoyi/ETDataset} comprises data originally collected for Informer~\citep{zhou2021informer}, including load and oil temperature measurements recorded at $15$-minute intervals between July 2016 and July 2018. The ETTh1 and ETTh2 subsets are sampled at $1$-hour intervals, while ETTm1 and ETTm2 are sampled at $15$-minute intervals.
    \item The Electricity dataset\footnote{https://archive.ics.uci.edu/ml/datasets/ElectricityLoadDiagrams20112014} contains hourly electricity consumption data for $321$ customers from 2012 to 2014.
    \item The Traffic dataset\footnote{http://pems.dot.ca.gov} consists of hourly road occupancy rates, collected by various sensors deployed on freeways in the San Francisco Bay area, sourced from the California Department of Transportation.
    \item The Weather dataset\footnote{https://www.bgc-jena.mpg.de/wetter/} includes local climatological data from nearly $1,600$ locations across the United States, covering a period of four years (2010 to 2013), with data points recorded at $1$-hour intervals.
\end{itemize}

\subsection{Baseline Models}\label{sec:baselines}
Here is a brief description of the baseline models used in this paper.

\begin{itemize}
    \item FEDformer~\citep{zhou2022fedformer} is a Transformer-based model proposing seasonal-trend decomposition and
exploiting the sparsity of time series in the frequency domain. The source code is available at \url{https://github.com/DAMO-DI-ML/ICML2022-FEDformer}.
    \item TimesNet~\citep{wu2022timesnet} is a CNN-based model with TimesBlock as a task-general backbone. It transforms 1D time series into 2D tensors to capture intraperiod and interperiod variations. The source code is available at \url{https://github.com/thuml/TimesNet}.
    \item TimeMixer~\citep{wang2024timemixer} is a fully MLP-based architecture with PDM and FMM blocks to take full advantage of disentangled multiscale series in both past extraction and future prediction phases. The source code is available at \url{https://github.com/kwuking/TimeMixer}.
    \item PatchTST~\citep{nie2022time} is a transformer-based model utilizing patching and CI technique. It also enables effective pre-training and transfer learning across datasets. The source code is available at \url{https://github.com/yuqinie98/PatchTST}.
    \item DLinear~\citep{zeng2023transformers} is an MLP-based model with just one linear layer, which outperforms Transformer-based models in LTSF tasks. The source code is available at \url{https://github.com/cure-lab/LTSF-Linear}.
    \item FITS~\citep{xu2023fits} is a linear model that manipulates time series data through interpolation in the complex frequency domain. The source code is available at \url{https://github.com/VEWOXIC/FITS}.
    \item SparseTSF~\citep{lin2024sparsetsf} a novel, extremely lightweight model for LTSF, designed to address the challenges of modeling complex temporal dependencies over extended horizons with minimal computational resources. The source code is available at \url{https://github.com/lss-1138/SparseTSF}.
\end{itemize}

\section{Advantages of MMFT}\label{sec:mmft-sft}

MMFT leverages a multi-scale approach to address the limitations of SFT. By operating across multiple scales, MMFT offers several key features:
\begin{itemize}
    \item \textbf{Good Adaptability to Non-Stationarity}. Non-stationarity in time series, where statistical properties such as trends or seasonality evolve over time, presents a challenge for SFT, which assumes stationarity. MMFT mitigates this limitation by decomposing the time series into multiple frequency components, each of which captures specific temporal patterns (e.g., short-term fluctuations or long-term trends). By adapting to non-stationary characteristics that SFT may overlook, MMFT effectively reduces bias. For instance, in financial datasets with shifting trends, MMFT can simultaneously analyze long-term patterns and short-term variations, enhancing predictive performance.
    \item \textbf{Effectively Capturing of Local and Global Patterns}. Time series data often contain both short-term (local) and long-term (global) patterns. SFT’s reliance on a single global scale may fail to capture local variations, leading to higher prediction variance. MMFT addresses this issue by operating at multiple scales, allowing it to capture local patterns at finer resolutions and global trends at coarser ones. This enables MMFT to better adapt to the varying characteristics of the data, reducing both overfitting to local noise and underfitting of broader trends. For example, MMFT can effectively model daily temperature fluctuations alongside longer seasonal cycles, leading to improved predictive accuracy across different time horizons.
    \item \textbf{Employing Learnable Frequency Masks}. MMFT introduces learnable frequency masks that selectively filter out irrelevant frequency components while retaining the key frequencies necessary for accurate prediction. Unlike the static filters used in SFT, these masks are optimized during training, enabling the model to focus on informative frequency components while discarding noise. This adaptive filtering process reduces both bias and variance, further enhancing model performance.
\end{itemize}

\section{More Analysis on Mask Output}\label{sec:mask_output}

\begin{figure}[h]
\centering

\begin{minipage}{0.32\textwidth}
    \centering
    \includegraphics[width=\textwidth]{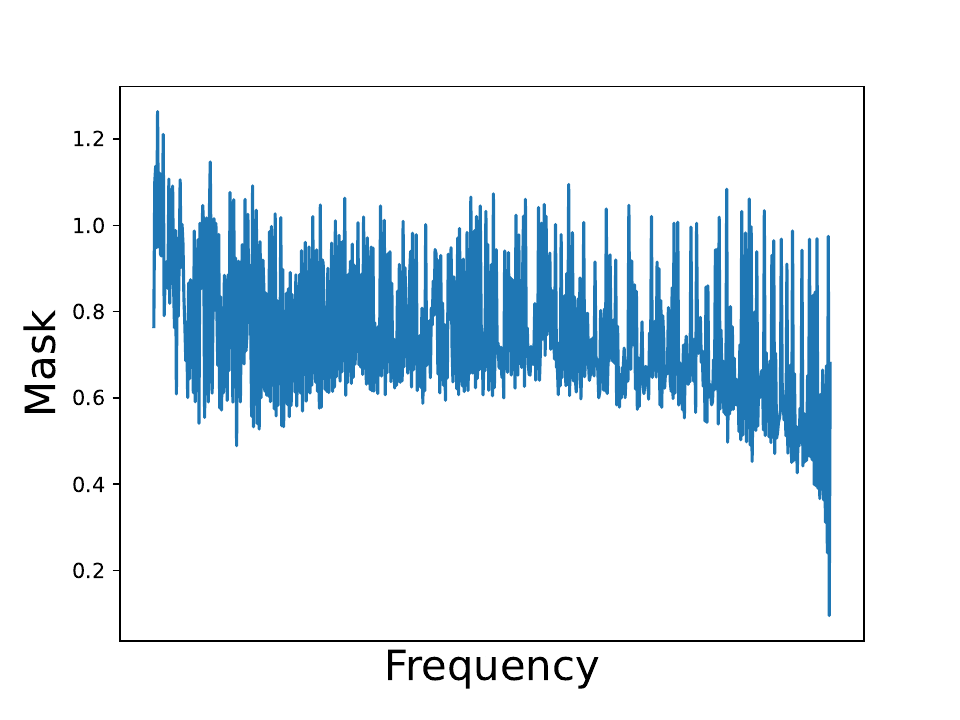}
    \caption*{Fine-scale}
    \label{fig:ETTh1_fine_scale}
\end{minipage}
\hfill
\begin{minipage}{0.32\textwidth}
    \centering
    \includegraphics[width=\textwidth]{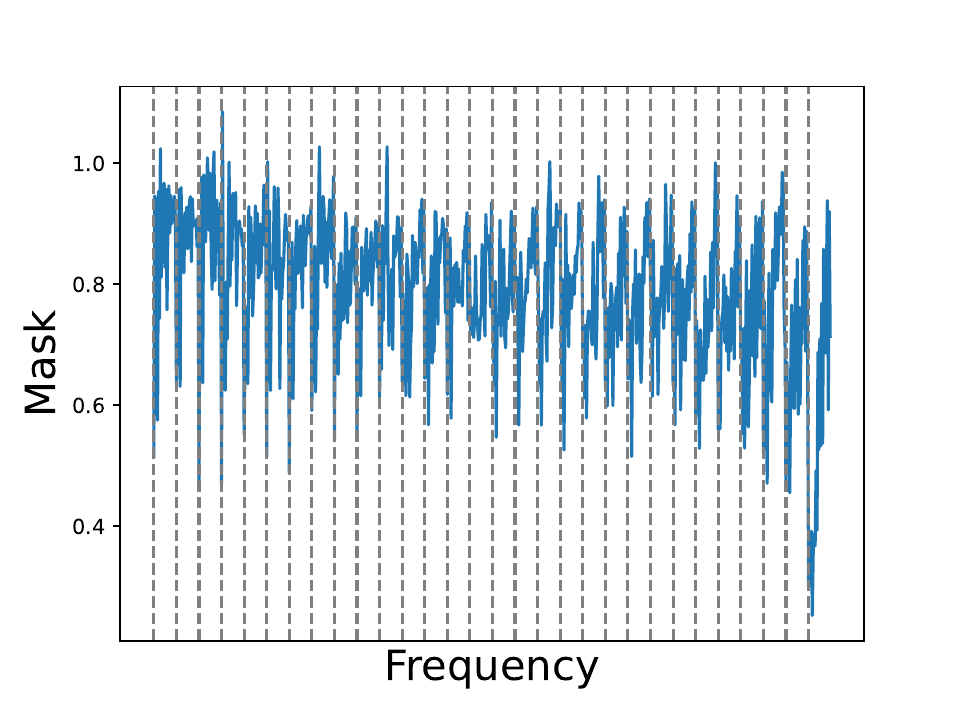}
    \caption*{Intermediate-scale }
    \label{fig:ETTh1_intermediate_scale}
\end{minipage}
\hfill
\begin{minipage}{0.32\textwidth}
    \centering
    \includegraphics[width=\textwidth]{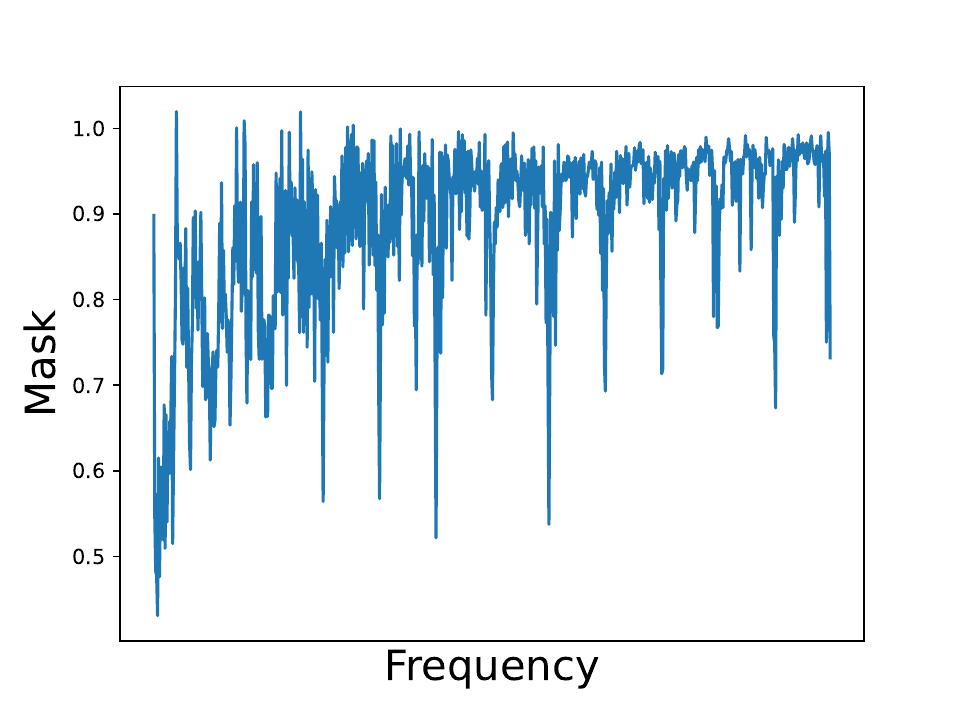}
    \caption*{Coarse-scale}
    \label{fig:ETTh1_coarse_scale}
\end{minipage}

\vspace{1em} 

\caption{Mask outputs for different frequency decompositions on the ETTh1 dataset. The segment lengths for the fine-scale, intermediate-scale, and coarse-scale decompositions are set to $2$, $24$, and $720$, respectively.}
\label{fig:mask_outputs_etth1}
\end{figure}

\begin{figure}[h]
\centering

\begin{minipage}{0.32\textwidth}
    \centering
    \includegraphics[width=\textwidth]{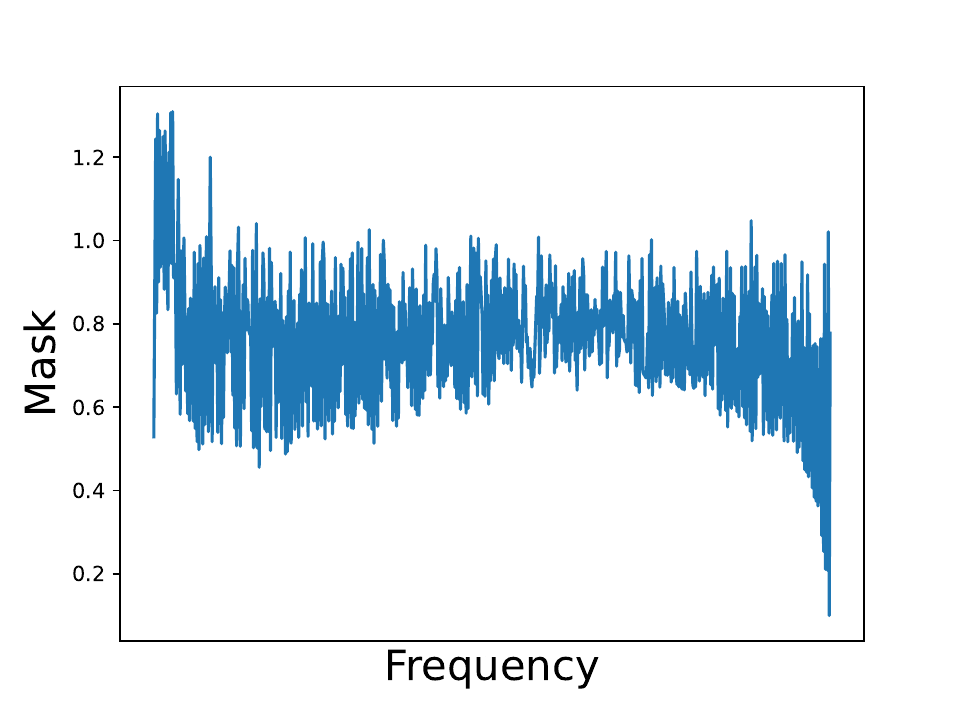}
    \caption*{Fine-scale}
    \label{fig:ETTh2_fine_scale}
\end{minipage}
\hfill
\begin{minipage}{0.32\textwidth}
    \centering
    \includegraphics[width=\textwidth]{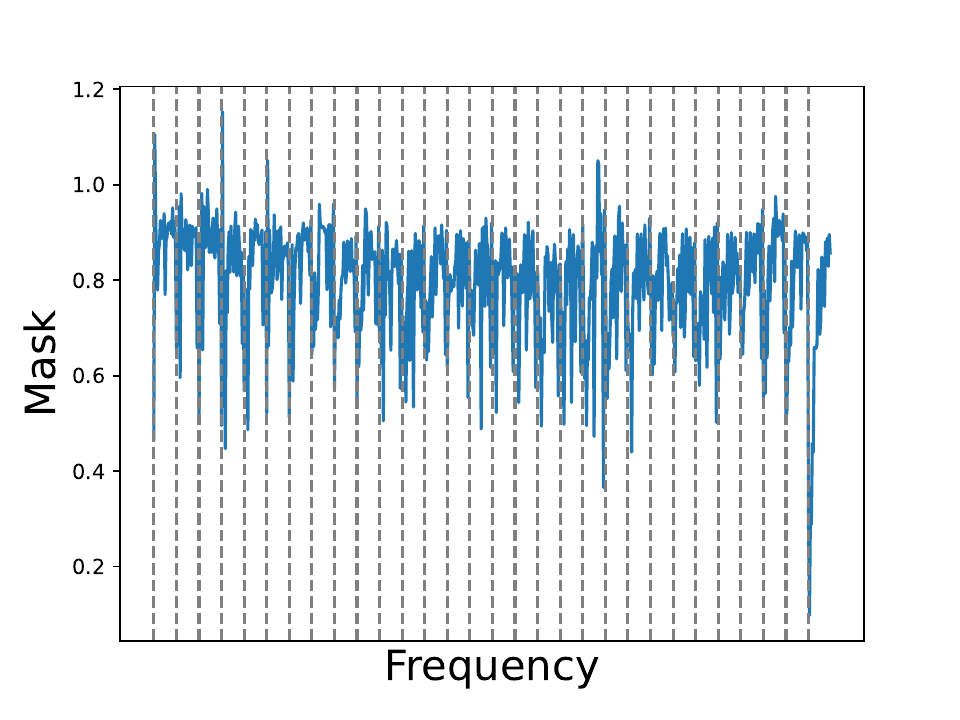}
    \caption*{Intermediate-scale }
    \label{fig:ETTh2_intermediate_scale}
\end{minipage}
\hfill
\begin{minipage}{0.32\textwidth}
    \centering
    \includegraphics[width=\textwidth]{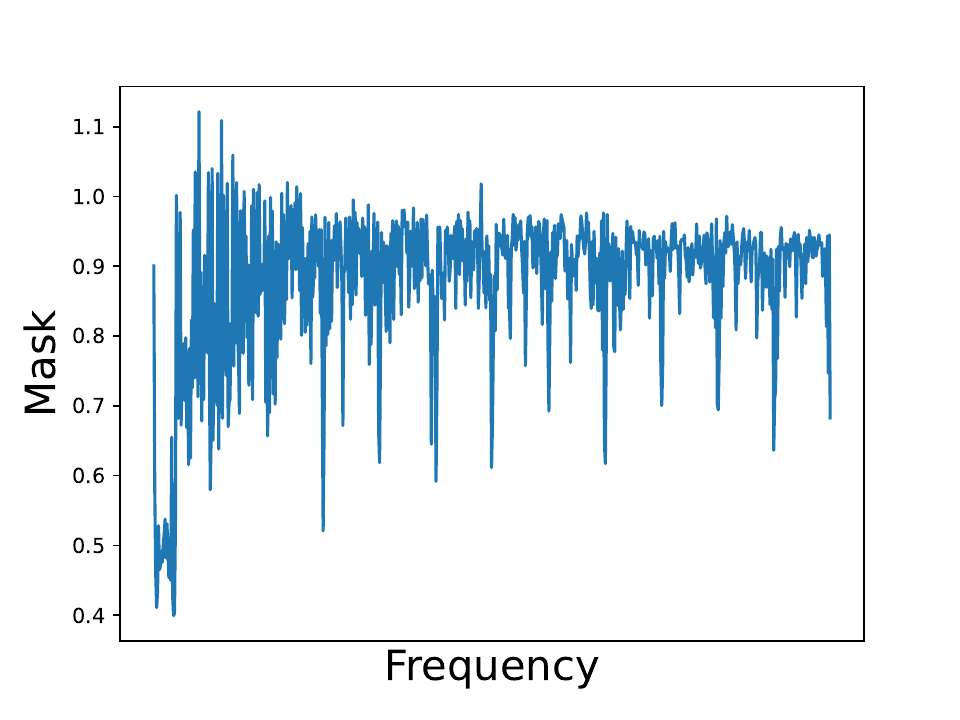}
    \caption*{Coarse-scale}
    \label{fig:ETTh2_coarse_scale}
\end{minipage}
\vspace{1em} 

\caption{Mask outputs for different frequency decompositions on the ETTh2 dataset. The segment lengths for the fine-scale, intermediate-scale, and coarse-scale decompositions are set to $2$, $24$, and $720$, respectively.}
\label{fig:mask_outputs_etth2}
\end{figure}

To analyze the learned masks at different scales, we visualize the mask outputs. Figures~\ref{fig:mask_outputs_etth1} and~\ref{fig:mask_outputs_etth2} illustrate the mask outputs for various frequency decompositions applied to the ETTh1 and ETTh2 datasets. These figures reveal a consistent pattern across time segments and show that the masks predominantly target high-frequency components with larger mask values indicating more aggressive attenuation of these components compared to lower-frequency ones.

A more detailed examination shows that the degree of frequency attenuation varies across time segments and scales. Specifically, in the fine-scale and intermediate-scale scenarios, high-frequency components exhibit a greater attenuation ratio in the earlier time segments. This observation suggests that at finer temporal resolutions, earlier time points experience a more significant reduction in high-frequency information. In contrast, for the coarse-scale scenario, the most recent time segments display a higher attenuation ratio for high-frequency components.

This pattern suggests that, as the temporal resolution increases, the model increasingly focuses on masking high-frequency components more aggressively in earlier time segments. Conversely, at coarser scales, more recent data points are more heavily filtered. This behavior likely reflects the model's adaptive strategy to prioritize different temporal patterns or noise levels depending on the scale of the analysis. The variation in attenuation ratios across scales and time segments indicates a nuanced approach to frequency masking, which may optimize the model's performance by selectively emphasizing or de-emphasizing specific temporal features based on their relevance at each scale.

The consistency of the Mask outputs across these scales suggests that the frequency decomposition method is both robust and effective in isolating different aspects of the time series data. Fine-scale outputs are particularly useful for identifying rapid fluctuations and short-term patterns, while coarse-scale outputs are essential for understanding broader trends and long-term behavior in the data. This multi-scale approach is highly beneficial for time series forecasting, as it allows the model to leverage both the fine details and the overarching trends, leading to more accurate and comprehensive predictions.

\end{document}